\documentclass[runningheads]{llncs}

\usepackage{eccv}

\usepackage{algorithmic}
\usepackage{amssymb}
\usepackage{amstext}
\usepackage{amsmath}
\usepackage{booktabs}
\usepackage[ruled,vlined]{algorithm2e}
\usepackage{xcolor}

\usepackage{multirow}
\usepackage{graphicx}
\usepackage{subcaption}
\usepackage{xcolor}

\usepackage{eccvabbrv}

\usepackage{graphicx}
\usepackage{booktabs}

\usepackage{caption}
\usepackage{subcaption}

\usepackage[accsupp]{axessibility}  

\usepackage[pagebackref,breaklinks,colorlinks,citecolor=eccvblue]{hyperref}

\usepackage{orcidlink}

\begin{document}

\title{DSMix: Distortion-Induced Sensitivity Map Based Pre-training for No-Reference Image Quality Assessment} 

\titlerunning{DSMix: Distortion-Induced Sensitivity Map Based Pre-training for NR-IQA}

\author{Jinsong Shi\inst{1,2}\orcidlink{0000-0002-7471-8221} \and
Pan Gao \inst{1,2}\thanks{Corresponding author}\orcidlink{0000-0002-4492-5430} \and
Xiaojiang Peng\inst{3}\orcidlink{0000-0002-5783-321X} \and Jie Qin \inst{1,2}\orcidlink{0000-0002-0306-534X}
}

\authorrunning{Shi~et al.}

\institute{College of Artificial Intelligence, Nanjing University of Aeronautics and Astronautics \\
\and
The Key Laboratory of Brain-Machine Intelligence
Technology, Ministry of Education, Nanjing, 211106, China\\
\and
College of Big Data and Internet, Shenzhen Technology University\\
\email{\{srache,pan.gao\}@nuaa.edu.cn, {pengxiaojiang}@sztu.edu.cn, {qinjiebuaa}@gmail.com}}


\maketitle

\begin{abstract}
Image quality assessment (IQA) has long been a fundamental challenge in image understanding. In recent years, deep learning-based IQA methods have shown promising performance. 
However, the lack of large amounts of labeled data in the IQA field has hindered further advancements in these methods. This paper introduces DSMix, a novel data augmentation technique specifically designed for IQA tasks, aiming to overcome this limitation. DSMix leverages the distortion-induced sensitivity map (DSM) of an image as prior knowledge. It applies cut and mix operations to diverse categories of synthetic distorted images, assigning confidence scores to class labels based on the aforementioned prior knowledge. In the pre-training phase using DSMix-augmented data, knowledge distillation is employed to enhance the model's ability to extract semantic features. 
Experimental results on both synthetic and authentic IQA datasets demonstrate the significant predictive and generalization performance achieved by DSMix, without requiring fine-tuning of the full model. 
Code is available at \url{https://github.com/I2-Multimedia-Lab/DSMix}.
\end{abstract}

\section{Introduction}
With the rise of the Internet, a large number of images are generated daily on social media platforms like Twitter and Instagram. However, the quality of these images can be compromised during acquisition, transmission, and compression processes. To ensure a good user experience, it is vital to develop effective methods for Image Quality Assessment (IQA).
The objective evaluation methods can be classified into three categories: Full-Reference (FR)~\cite{wang2004image, zhang2011fsim, zhang2018unreasonable}, Reduced-Reference (RR)~\cite{rehman2012reduced}, and No-Reference methods (NR)~\cite{mittal2012no}. FR-IQA methods compare distorted images to their reference images to assess the quality, while RR-IQA methods use partial information from the reference image. NR-IQA methods evaluate the quality of an image without any information from the reference image. Since obtaining a reference image for a distorted image in practical IQA scenarios is often challenging, researchers have focused on developing NR-IQA methods.

\begin{figure}[]
	\centering
	\includegraphics[width=0.85\textwidth]{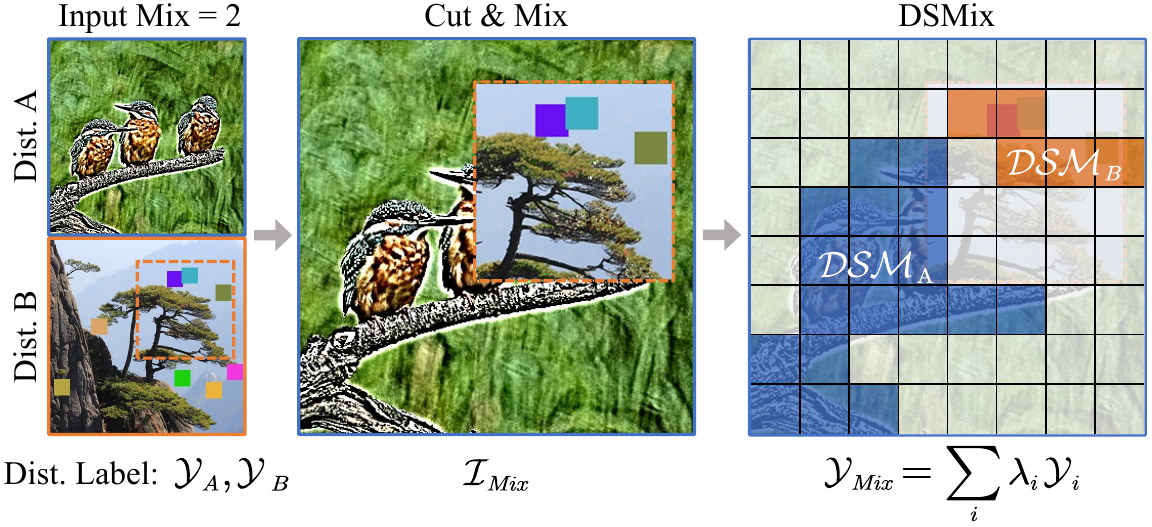}
	\caption{An overview of DSMix method. Dist. A and Dist. B denote distorted images with corresponding distortion type labels $\mathcal{Y}_A$ and $\mathcal{Y}_B$, respectively. By performing cut and mix operations on these two types of distorted images, the mixed image $\mathcal{I}_{Mix}$ is generated with a corresponding label $\mathcal{Y}_{Mix}$. The value of $\lambda$ is obtained by calculating the area ratio of $\mathcal{DSM}_A$ (\textcolor{blue}{blue area}) and $\mathcal{DSM}_B$ (\textcolor{orange}{orange area}) with respect to the entire DSM.}
	\label{fig:pre}
\end{figure}

Recently, deep learning-based NR-IQA methods have made significant progress on some synthetic and real datasets. However, this is far from enough, especially on authentic datasets, where SRCC has not exceeded 0.7 on the FLIVE dataset~\cite{ying2020patches,golestaneh2022no,saha2023re,zhao2023quality}. This is due to that these methods are still limited by the lack of labeled data. The FLIVE dataset only contains less than 40,000 images, while in the field of deep learning, the basic data volume is in the millions. Therefore, the existing data volume is insufficient to support NR-IQA tasks.
To tackle this dilemma, some NR-IQA methods~\cite{ma2017end,ke2021musiq,golestaneh2022no} adopt the strategy of dividing the image into multiple patches during training, with each patch's quality being derived from the mean opinion score (MOS) or differential mean opinion score (DMOS) value of the original distorted image. However, this approach is evidently flawed as the quality of image patches has a non-linear relationship with the overall image quality in the IQA field~\cite{ying2020patches}. Furthermore, some methods~\cite{zhang2018blind,ke2021musiq} attempt to perform pre-training for classification on ImageNet~\cite{deng2009imagenet}, followed by fine-tuning on specific IQA datasets. Nevertheless, the pre-training task on ImageNet is not specifically designed for IQA, and even with pre-training, it cannot recognize well quality degradation induced in images.

More recently, self-supervised NR-IQA methods~\cite{madhusudana2022image,zhao2023quality,saha2023re} have been gaining popularity. \cite{madhusudana2022image} develops a contrastive learning-based training framework with distortion type and level as auxiliary classification tasks, aiming to obtain more efficient image quality representations. \cite{zhao2023quality} and~\cite{saha2023re} adopt the unsupervised training framework proposed by MoCo-v2~\cite{chen2020improved} and learn the features of distortion and content in images by constructing their respective distortion degradation prediction schemes. However, these self-supervised methods all face the same obstacle - the inability to quantify the degree of distortion during training, which ultimately leads to their model performance reaching a bottleneck in the case of unavailability of reference image. 

In contrast to previous methods, we propose a new data augmentation method specifically designed for IQA tasks for self-supervised pre-training.
Taking two images with different distortion types, as shown in \cref{fig:pre}, as an example, we aim to enhance the given distorted image $A$ by randomly mixing it with a patch of arbitrary size cropped from distorted image $B$ at a random position in image $A$. The resulting mixed image is assigned a class label based on their respective distortion-induced sensitivity map ($\mathcal{DSM}$) of the corresponding distortion types, which is specifically weighted by the confidence score of the distortion type label for each image, denoted as $\lambda$.
By mixing images using the DSMix method, it is possible to obtain a larger and more diverse set of labeled data. Furthermore, the use of $\mathcal{DSM}$ to assign weights to the distortion types in the labeling process addresses the limitation of previous self-supervised methods, which were unable to quantitatively measure the degree of distortion.
Moreover, the image content itself is closely linked to perceptual quality as demonstrated in prior studies~\cite{li2018has,sun2023blind}. To imbue semantic representation into the model during the self-supervised pre-training of DSMix, we conduct knowledge distillation from a classification model trained on ImageNet. Importantly, no fine-tuning is performed, and only linear probing is utilized to derive the quality score. Consequently, experimental results across seven benchmark datasets showcase that our proposed method attains state-of-the-art (SOTA) performance at a substantially  reduced training expense.

\section{Related Work}
Traditional NR-IQA methods mainly relied on natural scene statistics (NSS), which assume that the visual features of images with intact quality follow certain statistical distribution rules, and different types and levels of distortion would perturb such distribution.
Based on this theory, methods have been developed using spatial domain~\cite{mittal2012no,mittal2012making}, gradients~\cite{zhang2015feature}, and discrete cosine transform (DST)~\cite{saad2012blind}. In addition, there are also some learning-based methods, such as constructing codebooks~\cite{ye2012unsupervised,xu2016blind} for quality feature representation. Although these methods performed well on some synthesized distortion datasets, they are unable to simulate real-world distortions, resulting in poor performance on authentic datasets.

With the rise of deep learning, early NR-IQA methods~\cite{zeng2017probabilistic,ma2017end,bosse2017deep,zhang2018blind,ying2020patches,zhu2020metaiqa,su2020blindly} used CNN backbones trained for image classification on ImageNet to extract features, and then regressed the features to the quality scores of distorted images. These deep learning methods have to some extent addressed the poor performance of traditional NR-IQA methods on authentic datasets. \cite{zhang2018blind} proposed a dual-branch structure, which conducts pre-training on both synthetic and authentic datasets, and finally uses bilinear pooling for feature fusion of the two branches. \cite{zhu2020metaiqa} proposes to learn meta-knowledge shared by different distortion types of images using meta-learning. \cite{su2020blindly} extracts content features of different scales from the model and pools them to predict the image quality. 
With the significant breakthrough of ViT~\cite{dosovitskiy2020image} in the field of computer vision, some Transformer-based NR-IQA methods are also gaining popularity. \cite{ke2021musiq} proposes hash-based 2D spatial embedding to evaluate images of arbitrary resolution. \cite{golestaneh2022no} proposes a hybrid Transformer structure that combines relative ranking and self-consistency. 

However, in the domain of NR-IQA, these methods are significantly constrained by the lack of labeled data, with Transformer-based methods particularly susceptible to overfitting.
Currently, some research efforts are devoted to addressing this obstacle. \cite{ying2020patches} attempts to establish the relationship between image patches and the entire image, and build a dataset of 120,000 image patches containing labeled data. 
While this approach is viable, there is a risk of sampling inaccurately or gathering patches that do not adequately represent the quality during the collection process.
Meanwhile, some self-supervised methods have also emerged. \cite{madhusudana2022image} divides images with different types and levels of distortion into different classes, and uses contrastive learning to distinguish synthetic distorted images and user-generated content (UGC). However, these categories actually belong to distortion labels, which are irrelevant to the quality of the images themselves. \cite{zhao2023quality} and \cite{saha2023re} use the MoCo-v2 framework for self-supervised pre-training of content and degradation on ImageNet. However, these methods cannot quantify the degree of distortion in the pre-training process, and still unable to assess the quality of labeled distorted images. In addition, their training costs are  extremely high and unaffordable, with \cite{zhao2023quality} using 8 NVIDIA V100 GPUs and \cite{saha2023re} even using 18 NVIDIA A100-40GB GPUs, which is undoubtedly unfriendly to NR-IQA task.

\begin{figure*}[ht]
	\centering
	\includegraphics[width=0.995\textwidth]{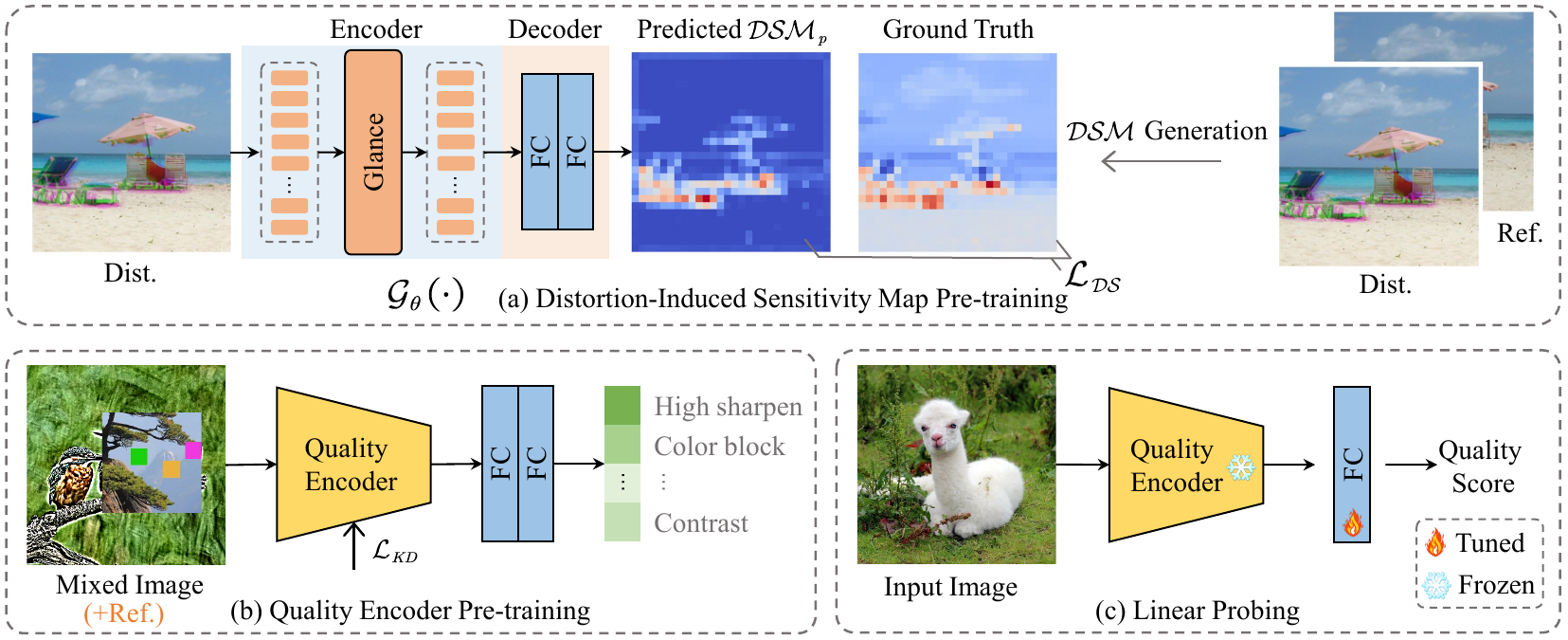}
	\caption{An overview of our proposed NR-IQA method based on DSMix. (a) The generation of DSM. (b) Self-supervised pre-training based on DSMix. (c) Linear probing on the specific IQA dataset.}
	\label{fig:model}
\end{figure*}

In contrast, our proposed DSM-based data augmentation method is simple to implement, and can generate more diverse distorted image data for training. All experiments can be completed on only a single NVIDIA 3090 GPU.

\section{Method}
\label{sec:method}
In this section, we will provide a detailed explanation of the proposed DSMix framework, as depicted in \cref{fig:model}.

\subsection{Distortion-induced Sensitivity Map Pre-training}
\begin{figure}[h]
	\centering
	\includegraphics[width=0.7\textwidth]{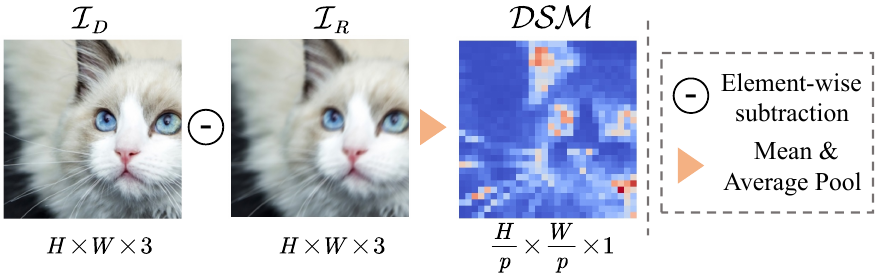}
	\caption{Overview of GT DSM generation process.}
	\label{fig:dsm}
\end{figure}

\noindent
\textbf{Ground-truth (GT) DSM Generation.}
Given a distorted image, denoted as $\mathcal{I}_D \in \mathbb{R}^{H\times W \times 3}$ and its corresponding reference image denoted as $\mathcal{I}_R \in \mathbb{R}^{H\times W \times 3}$, where $H$ and $W$ represent the height and width respectively. As illustrated in \cref{fig:dsm}, the first step is to compute the absolute difference between $\mathcal{I}_D$ and $\mathcal{I}_R$. This difference is then averaged over the channels, resulting in $\mathcal{I}_{Diff}$. Subsequently, $\mathcal{I}_{Diff}$ undergoes average pooling with a kernel size of $p$, yielding the GT $\mathcal{DSM}$ representation.

\noindent
\textbf{DSM pre-training.}
In the IQA domain, considering the distortion sensitivity of an image mainly relies on its global information, we adopt the Vision Transformer (ViT) architecture as the encoder. Specifically, we remove the CLS token from the original ViT, while retaining the position embeddings ($PE$) and Transformer encoder layers, denoted as $EN_i$, where $i\in\{1,2,...,L\}$. Given the input $\mathcal{I}_D$, we divide it into patches of size $p$ and linearly project them to $D$ dimension, denoted as $\mathcal{I}_{proj}\in \mathbb{R}^{\frac{H}{p}\times\frac{W}{p}\times D }$. Then, $\mathcal{I}_{proj}$ is added with $PE$ and passed through $L$ layers of the $EN$ module. As for the decoder part, we first apply an MLP layer to the output of $EN_L$, reshape it, and obtain the final predicted $\mathcal{DSM}_p\in \mathbb{R}^{\frac{H}{p}\times\frac{W}{p}\times 1} $, which can be formulated below:
\begin{equation}
	\underset{\mathcal{G}_\theta(\cdot )}{\underbrace{\text{MLP}\left( EN_L\sim EN_1\left( \mathcal{I}_{proj}+PE \right) \right)}} \rightarrow \mathcal{D}\mathcal{S}\mathcal{M}_p
\end{equation}
where $\to$ denotes the reshape operation. Let $\mathcal{G}_\theta(\cdot)$ represent the entire network with learnable parameters $\theta$. In the subsequent sections, $\mathcal{G}_\theta(\cdot )$ is assumed to include the reshape operation by default.
Mean squared error (MSE) is used during the pre-training process, which can be noted as:
\begin{equation}
	\mathcal{L_{DS}}=\frac{1}{HW}\sum_{i=1}^H{\sum_{j=1}^W{\left( \mathcal{DSM}_p\left( i,j \right) -\mathcal{DSM}\left( i,j \right) \right) ^2}}
\end{equation}

\subsection{Quality Encoder Pre-training}

\noindent
\textbf{DSMix.}
To create a wider range of inputs, we consider images with various types and levels of distortions as separate classes for mixed augmentation.
Let ($\mathcal{I}_{D_A}$, $\mathcal{Y}_A$) represent a distorted image pair with distortion $A$ and label $A$, and consider the mixing of this pair with another distorted image pair ($\mathcal{I}_{D_B}$, $\mathcal{Y}_B$) as an example (\ie, Mix=2). Specifically, we uniformly sample a patch $P(p_x, p_y,p_w,p_h)$ from $\mathcal{I}_{D_B}$ and replace the corresponding region in $\mathcal{I}_{D_A}$ with $P$. The resulting mixed image is denoted as $\mathcal{I}_{Mix}$:
\begin{align}
	\begin{aligned}
		& \mathcal{I}_{Mix}=\mathbf{M}\odot \mathcal{I}_{D_A}+(\mathbf{1}-\mathbf{M})\odot \mathcal{I}_{D_B} \\
		& \mathcal{Y}_{Mix}=\lambda \mathcal{Y}_A +(1-\lambda) \mathcal{Y}_B
	\end{aligned}
\end{align}
where $\mathbf{M}\in \{0,1\}^{HW}$ denotes a binary mask indicating where to drop out and fill in from two distorted images. $\mathbf{1}$ is a binary mask filled with ones, and $\odot$ is element-wise multiplication. $\lambda$ is the portion of $\mathcal{Y}_A$ in the mixed label, which is calculated as follows:
\begin{equation}
	\mathcal{DSM}^{'}=\left[ \mathcal{G}_{\theta}(\mathcal{I}_{Mix}) \right]_{\mathbf{M}} \cdot \uparrow 
\end{equation}
\begin{equation}
	\lambda =\frac{\sum (\mathcal{DSM}^{'}\cdot \mathbf{M})}{\sum \mathcal{DSM}^{'}}
\end{equation}
where $[~]_{\mathbf{M}}\cdot \uparrow$ denotes upsampling $\mathcal{DSM}\in \mathbb{R}^{\frac{H}{p}\times\frac{W}{p}\times 1}$ to the size of $H\times W$ using bilinear interpolation. $\lambda$ is the proportion of area attributed to $\mathcal{DSM}^{'}$ masked by $\mathbf{M}$.
In this way, the network can dynamically allocate weights to labels based on the distortion in the $\mathcal{DSM}^{'}$. 

\noindent
\textbf{Quality Encoder Pre-training (QEP).}
By leveraging the DSMix data augmentation method, we can generate a wide range of distorted images with their corresponding mixed labels. As illustrated in \cref{fig:model}.(b), the pre-training process involves utilizing the ResNet-50~\cite{he2016deep} backbone as the Quality Encoder (QE), along with two MLP heads for classifying distortion types and levels. Throughout the entire training process, we utilize a soft target form of cross-entropy loss $\mathcal{L}_{QC}$, which is defined as follows:
\begin{equation}
	\mathcal{L}_{QC} = - \frac{1}{N} \sum_{i=1}^{N} \sum_{j=1}^{C} \mathcal{Y}_{Mix}[i,j] \text{log}(\text{Softmax}(\hat{\mathcal{Y}}_{Mix}[i,j]))
\end{equation}
where $N$ is the number of images in the batch, $C$ is the number of classes, $\mathcal{Y}_{Mix}$ denotes the ground-truth labels in the soft target form of one-hot vectors, and $\hat{\mathcal{Y}}_{Mix}$ represents the mixed labels predicted by the model.

\noindent
\textbf{Semantic Knowledge Distillation.}
\begin{figure}[h]
	\centering
	\includegraphics[width=1.0\textwidth]{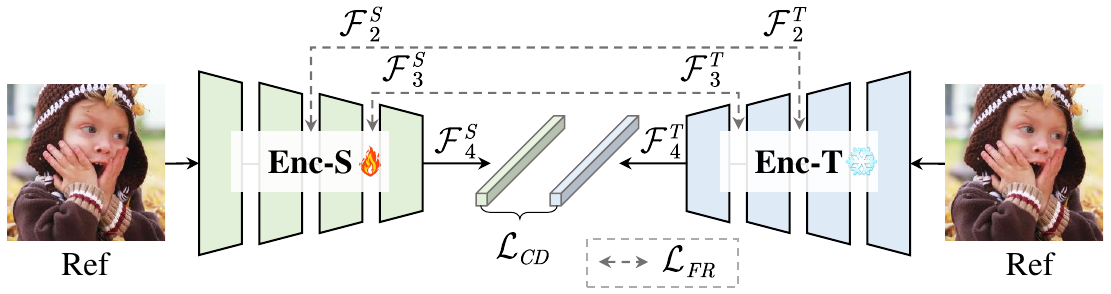}
	\caption{Overview of semantic knowledge distillation, where the flame and snowflake icons denote parameters frozen and tuned, respectively. Enc-S denotes QE, and Enc-T denotes teacher model.}
	\label{fig:kd}
\end{figure}
In the domain of IQA, an image's quality is closely tied to its semantic content, especially when evaluating real-world images with unknown diverse distortions. However, DSMix-based models lack prior knowledge about semantic information, hindering their performance. To overcome this limitation, in \cref{fig:kd}.(b), we propose a solution using ResNet-50 pre-trained on ImageNet as a teacher model and the DSMix-trained model as a student model. By employing knowledge distillation, we transfer the semantic knowledge from the teacher model to the student model, enabling it to better evaluate image quality by incorporating rich semantic insights. 
The knowledge distillation loss $\mathcal{L}_{KD}$ can be denoted as:
\begin{align}
	\begin{aligned}
		& \mathcal{L}_{KD} = \lambda_1\mathcal{L}_{FR} - \lambda_2\mathcal{L}_{CD} \\
		&\begin{cases}
			\mathcal{L}_{FR}=\sum_{l=2,3}{(\text{MAE}( \mathcal{F}_{l}^{S},\mathcal{F}_{l}^{T} ))}\\
			\mathcal{L}_{CD}=\text{CosD}(\text{AvgPool}( \mathcal{F}_{4}^{S}) ,\text{AvgPool}( \mathcal{F}_{4}^{T}))
		\end{cases}
	\end{aligned} 
\end{align}
where the $\mathcal{L}_{FR}$ is the feature reconstruction loss, and $\mathcal{L}_{CD}$ is the cosine distance loss, $\lambda_1$ and $\lambda_2$ are balancing coefficients. MAE stands for mean absolute error. $\mathcal{F}_{l}^{S}$ and $\mathcal{F}_{l}^{T}$ denote the feature maps of the student and teacher models at the $\mathcal{L}^{th}$ stage of ResNet-50, respectively. CosD refers to the cosine distance. By utilizing $\mathcal{L}_{KD}$, the student model can gain a better understanding and representation of the semantic in images at various levels of abstraction.

The DSMix-based QEP does not rely on subjective quality scores of distorted images as supervisory information. 
The weights of both model $\mathcal{G}_\theta(\cdot )$ and the teacher model remain frozen throughout the training process. 
The total loss is given as:
\begin{equation}
	\mathcal{L}_{Quality} = \mathcal{L}_{QC} + \mathbb{I}_{\in Ref} (\mathcal{L}_{KD})
\end{equation}
It is worth noting that in order to avoid the potential semantic inconsistency in the mixed image, we only perform semantic knowledge distillation on the reference image.

\subsection{Pseudo-code}
Algorithm \ref{algo:dsm} provides the pseudo-code of DSMix-based QEP in a pytorch-like style. The clean pseudo-code shows that only few lines of code can boost the performance in the plug-and-play manner.
\definecolor{commentcolor}{RGB}{110,154,155}   
\newcommand{\PyComment}[1]{\ttfamily\textcolor{commentcolor}{\# #1}}  
\newcommand{\PyCode}[1]{\ttfamily\textcolor{black}{#1}} 

\begin{algorithm}[ht]
	\SetAlgoLined
	\scriptsize
	\Indm
	~~~\PyComment{H, W: the height and width of the input image} \\
	\hspace{0.7em}\PyComment{M: 0-initialized mask with shape (H,W)} \\
	\hspace{0.7em}\PyComment{upsample: upsample from (H/p)*(W/p) to (H*W)} \\
	\hspace{0.7em}\PyComment{(bx1, bx2, by1, by2): bounding box coordinate} \\
	\hspace{0.7em}\PyComment{QE: student model, Tea: teacher model} \\
	\hspace{0.7em}\PyCode{for (x, y) in loader:} \PyComment{load a minibatch with N pairs}\\
	\Indp   
	\hspace{0.7em}\PyCode{M[:,:,M==1 = x.flip(0)[:,:,M==1]} \PyComment{Cut \& Mix} \\ 
	\BlankLine
	\hspace{0.7em}\PyComment{DSM: (H, W), F: features maps of each stage} \\
	\hspace{0.7em}\PyComment{The weights of G\_theta and Tea are frozen } \\
	\hspace{0.7em}\PyCode{DSM = upsampe(G\_theta(x))} \\
	\hspace{0.7em}\PyCode{logits, F\_stu = QE(x)} \\
	\hspace{0.7em}\PyCode{F\_tea = Tea(x)} \\
	\BlankLine
	\hspace{0.7em}\PyComment{Mix labels with the DSM} \\
	\hspace{0.7em}\PyCode{lam = sum(DSM × M)/sum(DSM)} \\
	\hspace{0.7em}\PyCode{y = (1-lam) * y + lam * y.flip(0)} \\ 
	\BlankLine
	\hspace{0.7em}\PyComment{QC: CrossEntropy loss with soft targets} \\
	\hspace{0.7em}\PyComment{KD: Semantic knowledge distillation loss} \\
	\hspace{0.7em}\PyCode{L\_Quality = QC(logits, y) + KD(F\_stu, F\_tea)} \\
	\hspace{0.7em}\PyCode{L\_Quality.backward()} \\
	\Indm 
	\caption{Pseudocode of DSMix-based QEP in a PyTorch-like style.}
	\label{algo:dsm}
\end{algorithm}

\subsection{Linear Probing (LP)}
In our experiments (\cref{fig:model}.(c)), the QE trained via DSMix-based QEP is used as a distortion extractor. An MLP regression head is then employed, with frozen parameters for the QE model. Unlike existing NR-IQA methods, our proposed model only trains the regression head to map feature maps to final quality scores. The training process utilizes the Smooth L1-loss~\cite{girshick2015fast}, noted as $\mathcal{L}_{Score}$:
\begin{equation}
	\mathcal{L}_{Score} = \frac {1}{N} \sum_{i\in N}
	\begin{cases}
		0.5  (x_i - y_i)^2, & \text{if } |x_i-y_i| < 1 \\
		|x_i-y_i| - 0.5, & \text{otherwise}
	\end{cases}
\end{equation}
where $x_i$ is the predicted quality score of the $i^{th}$ image, and $y_i$ denotes the corresponding subjective quality score.

\section{Experiments}
\subsection{Datasets and Evaluation Criteria}
\noindent
\textbf{Datasets.}
Our method is evaluated on seven publicly available IQA datasets, including LIVE~\cite{sheikh2006statistical}, CSIQ~\cite{larson2010most}, TID2013~\cite{ponomarenko2015image}, KADID~\cite{lin2019kadid}, CLIVE~\cite{ghadiyaram2015massive}, KonIQ~\cite{hosu2020koniq}, and FLIVE~\cite{ying2020patches}.
LIVE consists of 799 images with five types of distortions, each corresponding to five levels of distortion. CSIQ contains 866 images and uses six types of distortions based on 30 reference images. TID2013 consists of 3000 images with 24 types of distortions. KADID is currently the largest annotated synthetic distortion dataset, consisting of 10125 images, including 81 reference images and 25 distortion levels. CLIVE contains 1162 images with various distortions captured by mobile devices. KonIQ includes 10073 images selected from YFCC-100M~\cite{thomee2016yfcc100m} dataset. FLIVE is the largest authentic dataset so far, comprising 39810 real distorted images with different resolutions, making it the most challenging dataset for NR-IQA tasks. 
In the QEP process, we leverage the synthesized dataset KADIS, which contains 700k distorted images. These images are created using Matlab from 140k reference images, incorporating 25 diverse types of distortions. 
Details refer to \cref{tab:dset}.

\begin{table}
	\centering
	\small
	\caption{Summary of IQA datasets}
	\scalebox{0.85}{
	\begin{tabular}{ccccc}
		\toprule[1.3pt]
		Dataset & Size  & Distortion & MOS/DMOS &  Range            \\
		\midrule
		LIVE    & 799   & Synthetic  & DMOS     & [0,100]  \\
		CSIQ    & 866   & Synthetic  & DMOS     & [0,1]    \\
		TID2013 & 3000  & Synthetic  & MOS      & [0,9]    \\
		KADID   & 10125 & Synthetic  & DMOS     & [1,5]    \\
		KADIS   & 700k  & Synthetic  & -        & -                      \\
		CLIVE   & 1162  & Authentic  & MOS      & [0,100]  \\
		KonIQ   & 10073 & Authentic  & MOS      & [1,5]    \\
		FLIVE   & 39810 & Authentic  & MOS      & [0,100] \\
		\bottomrule[1.3pt]
	\end{tabular}}
	\label{tab:dset}
\end{table}

\noindent
\textbf{Evaluation criteria.}
We select Spearman's Rank-order Correlation Coefficient (SRCC) and Pearson's Linear Correlation Coefficient (PLCC) as metrics to measure prediction accuracy and monotonicity, respectively. 
Both coefficients have a range of [0, 1], where a higher PLCC indicates greater accuracy and a higher SRCC represents more accurate ranking of sample quality.

\subsection{Implementation Details}
All of our experiments are conducted using PyTorch on a single NVIDIA RTX 3090 GPU. 

\noindent
\textbf{DSM pre-training.}
In this phase, we utilize the KADIS dataset and leverage the ViT-B/8 Transformer structure. During training, images are randomly cropped to a size of 224$\times$224. For generating the ground truth, pooling size $p$ is set to 8. We use AdamP optimizer with weight decay of 0.0001 and batch size of 40. The initial learning rate is set to 0.005 and decayed using a cosine annealing strategy. The training is performed for a total of 10 epochs, with $L=6$ and $D=768$.

\begin{table*}[h!]
	\centering
	\caption{Performance comparison of DSMix \emph{v.s.} SOTA NR-IQA methods on synthetically and authentically distorted datasets. The $\ast$ means missing corresponding results in the original paper. The best and second-best results are highlighted in \textbf{\textcolor{red}{red}} bold and \underline{\textcolor{blue}{blue}} underlined, respectively.}
	\label{tab:perform}
	\scalebox{0.8}{
	\begin{tabular}{ccccccccc} 
		\toprule[1.3pt]
		SRCC      & LIVE & CSIQ & TID2013 & KADID & CLIVE & KonIQ & FLIVE  & Training Paradigm\\ 
		\midrule
		PQR$\ast$\cite{zeng2017probabilistic} &0.965 &0.872 &0.740    &-      &0.857  &0.880  &-       & \multirow{8}{*}{\emph{Training form scratch}}\\
		ILNIQE\cite{zhang2015feature}    &0.902 &0.822 &0.521    &0.534  &0.508  &0.523  &0.294   \\
		BRISQUE\cite{mittal2012no}   &0.929 &0.812 &0.626    &0.528  &0.629  &0.681  &0.303   \\
		WaDIQaM\cite{bosse2017deep}   &0.960 &0.852 &0.835    &0.739  &0.682  &0.804  &0.455   \\
		HyperIQA\cite{su2020blindly}  &0.962 &0.923 &0.840    &0.852  &0.859  &0.906  &0.544   \\
		MUSIQ$\ast$\cite{ke2021musiq} &-   &-     &-        &-      &-      &0.916  &\textbf{\textcolor{red}{0.646}}   \\
		TReS\cite{golestaneh2022no}      &0.969 &0.922 &\underline{\textcolor{blue}{0.863}}    &0.859  &0.846  &0.915  &0.554   \\
            TOPIQ\cite{chen2023topiq}  &-   &-     &-        &-     
            &0.870 &\underline{\textcolor{blue}{0.926}} &0.633 \\
		\midrule
  		DBCNN\cite{zhang2018blind}     &0.968 &0.946 &0.816    &0.851  &0.869  &0.875  &0.545 & \multirow{3}{*}{\emph{Pre-training + Fine tuning}}  \\
		MetaIQA\cite{zhu2020metaiqa}   &0.960 &0.899 &0.856    &0.762  &0.835  &0.887  &0.540   \\
		QPT$\ast$\cite{zhao2023quality} &-     &-     &-        &-      &\textbf{\textcolor{red}{0.895}}  &\textbf{\textcolor{red}{0.927}}  &0.610  \\
		\midrule
		CONTRIQUE\cite{madhusudana2022image} &0.960 &0.942 &0.843    &\underline{\textcolor{blue}{0.934}}  &0.845  &0.894  &0.580   & \emph{Pre-training + Ridge regression}\\
		Re-IQA\cite{saha2023re}    &\underline{\textcolor{blue}{0.970}}&\underline{\textcolor{blue}{0.947}} &0.804    &0.872  &0.840  &0.914  &\underline{\textcolor{blue}{0.645}}  & \emph{Pre-training + Linear probing} \\ 
		\midrule
		\textbf{Ours}      &\textbf{\textcolor{red}{0.974}}      & \textbf{\textcolor{red}{0.957}}     &\textbf{\textcolor{red}{0.906}}         &\textbf{\textcolor{red}{0.943}}       
		&\underline{\textcolor{blue}{0.873}}       &0.915      &\textbf{\textcolor{red}{0.646}} &  \emph{Pre-training + Linear probing}       \\ 
		\cmidrule[1.3pt]{1-9}
		PLCC      & LIVE & CSIQ & TID2013 & KADID & CLIVE & KonIQ & FLIVE   &Training Paradigm\\ 
		\midrule
		PQR$\ast$\cite{zeng2017probabilistic} &\underline{\textcolor{blue}{0.971}} &0.901 &0.798    &-      &0.882  &0.884  &-   & \multirow{8}{*}{\emph{Training form scratch}}    \\
		ILNIQE\cite{zhang2015feature}    &0.906 &0.865 &0.648    &0.558  &0.508  &0.537  &0.332   \\
		BRISQUE\cite{mittal2012no}   &0.944 &0.748 &0.571    &0.567  &0.629  &0.685  &0.341   \\
		WaDIQaM\cite{bosse2017deep}   &0.955 &0.844 &0.855    &0.752  &0.671  &0.807  &0.467   \\
		HyperIQA\cite{su2020blindly}  &0.966 &0.942 &0.858    &0.845  &0.882  &0.917  &0.602   \\
		MUSIQ$\ast$\cite{ke2021musiq} &-   &-     &-        &-      &-      &0.928  &\textbf{\textcolor{red}{0.739}}   \\
		TReS\cite{golestaneh2022no}      &0.968 &0.942 &\underline{\textcolor{blue}{0.883}}    &0.858  &0.877  &0.928  &0.625   \\
              TOPIQ\cite{chen2023topiq}  &-   &-     &-        &-    
            &\underline{\textcolor{blue}{0.884}} &\underline{\textcolor{blue}{0.939}} &0.722 \\
		\midrule
  		DBCNN\cite{zhang2018blind}     &\underline{\textcolor{blue}{0.971}} &0.959 &0.865    &0.856  &0.869  &0.884  &0.551 & \multirow{3}{*}{\emph{Pre-training + Fine tuning}}  \\
		MetaIQA\cite{zhu2020metaiqa}   &0.959 &0.908 &0.868    &0.775  &0.802  &0.856  &0.507   \\
		QPT$\ast$\cite{zhao2023quality} &-     &-     &-        &-      &\textbf{\textcolor{red}{0.914}}  &\textbf{\textcolor{red}{0.941}}  &0.677  \\
		\midrule
		CONTRIQUE\cite{madhusudana2022image} &0.961 &0.955 &0.857    &\underline{\textcolor{blue}{0.937}}  &0.857  &0.906  &0.641  & \emph{Pre-training + Ridge regression} \\
		Re-IQA\cite{saha2023re}    &\underline{\textcolor{blue}{0.971}} &\underline{\textcolor{blue}{0.960}} &0.861    &0.885  &0.854  &0.923  &0.733  & \emph{Pre-training + Linear probing} \\ 
		\midrule
		\textbf{Ours}      &\textbf{\textcolor{red}{0.974}}      &\textbf{\textcolor{red}{0.962}}      &\textbf{\textcolor{red}{0.922}}         &\textbf{\textcolor{red}{0.945}}       
		&0.883       &0.925      &\underline{\textcolor{blue}{0.735}}    & \emph{Pre-training + Linear probing}    \\
		\bottomrule[1.3pt]
	\end{tabular}}
\end{table*}

\newcommand{\smallblack}[1]{\textbf{\textcolor{gray}{\scriptsize{#1}}}}
\begin{table*}[h!]
	\caption{Replacing the pre-trained weights of existing SOTA methods with DSMix leads to an improvement in performance. The experiments are carried out utilizing the open-source code, while ensuring consistent partitioning of the dataset and training hyperparameters.}
	\label{tab:replace}
	\centering
	\scalebox{0.75}{
	\begin{tabular}{@{}cc|ll|ll|ll@{}}
		\toprule[1.3pt]
		\multirow{2}{*}{Method}   & \multirow{2}{*}{\begin{tabular}[c]{@{}c@{}}Pre-trained \\ Type\end{tabular}} & \multicolumn{2}{c|}{CSIQ} & \multicolumn{2}{c|}{FLIVE} & \multicolumn{2}{c}{KADID} \\ 
		&                                                                              & SRCC        & PLCC       & SRCC        & PLCC        & SRCC        & PLCC        \\
		\cmidrule{1-8}
		\multirow{2}{*}{HyperIQA\cite{su2020blindly}} 
		& Original  & 0.923 & 0.942 & 0.544 & 0.602 & 0.852 & 0.845       \\
		& DSMix   & \textbf{0.939} \smallblack{+0.016} & \textbf{0.954} \smallblack{+0.012} 
		& \textbf{0.631} \smallblack{+0.087}      & \textbf{0.722} \smallblack{+0.120} 
		& \textbf{0.898} \smallblack{+0.046}      & \textbf{0.902} \smallblack{+0.057}       \\ \cmidrule{1-8}
		\multirow{2}{*}{TReS\cite{golestaneh2022no}}     
		& Original  & 0.922 & 0.942 & 0.554 & 0.625 & 0.859 & 0.858       \\
		& DSMix   & \textbf{0.941} \smallblack{+0.019} & \textbf{0.956} \smallblack{+0.014}  & \textbf{0.633} \smallblack{+0.079}      & \textbf{0.718} \smallblack{+0.093}   & \textbf{0.904} \smallblack{+0.045}      & \textbf{0.910} \smallblack{+0.052}            \\    
		\bottomrule[1.3pt]        
	\end{tabular}}
\end{table*}

\begin{table}[]
	\centering
	\caption{Cross dataset SRCC comparasion of NR-IQA models, where bold entries indicate the best performance. The ``-'' indicates missing data in the original paper.}
	\label{tab:cross}
	\scalebox{0.85}{
	\begin{tabular}{c|c|c|c|c|c|cc} 
		\toprule[1.3pt]
		Trained on & CLIVE & KonIQ & LIVE  & CSIQ  & LIVE    & FLIVE & FLIVE\\ 
		\cmidrule{1-8}
		Test on    & KonIQ & CLIVE & CSIQ  & LIVE  & TID2013 & CLIVE & KonIQ\\ 
		\midrule
		PQR\cite{zeng2017probabilistic} 
		& 0.757 & 0.770 & 0.719 & 0.922 & 0.548 & 0.716 & 0.744            \\
		Hyper-IQA\cite{su2020blindly}  
		& 0.772 & 0.785 & 0.744 & 0.926 & 0.551 & 0.735 & 0.758            \\
		CONTRIQUE\cite{madhusudana2022image}  
		& 0.676 & 0.731 & 0.823 & 0.925 & 0.564 & 0.698 & 0.703            \\
		Re-IQA\cite{saha2023re}     
		& 0.769 & \textbf{0.791}  & 0.808 & 0.929  & - & - &-              \\ 
		\cmidrule{1-8}
		Ours   
		& \textbf{0.790}   & \textbf{0.791}   & \textbf{0.842}   & \textbf{0.935}  & \textbf{0.583}   & \textbf{0.772}   & \textbf{0.784}             \\
		\bottomrule[1.3pt]
	\end{tabular}}
\end{table}

\noindent
\textbf{QEP.}
In this phase, we also training on the KADIS dataset, but with the addition of 140K reference images. When using DSMix data augmentation, the images are randomly horizontally flipped and then randomly cropped to a size of 224$\times$224. Mix=3, allowing for a maximum of mixing two different categories of distorted images. 
Specifically, during multi-class training, the KADIS dataset contains 25 types of distortions and 5 levels of distortion, resulting in a total of 126 classes including the reference images. 
Theoretically, the degradation space size of the 125 classes (reference images not using data augmentation) is ${\textstyle \sum_{125}^{i}} \textbf{C}_9^i \times \textbf{A}_i^i \approx 2\times10^6$.
We select ResNet-50 as the QE and add an MLP as the classification head. For semantic knowledge distillation on the reference images, we set $\lambda_1=0.1$ and $\lambda_2=1$. The teacher model chosen for distillation is a ResNet-50 pre-trained on ImageNet. We use the SGD optimizer with a weight decay of 0.0001, momentum of 0.9, and a batch size of 110. The initial learning rate is set to 0.01 and multiplied by 0.9 every five epochs. The training process consists of 120 epochs.

\noindent
\textbf{LP.}
In this step, we randomly apply horizontal flipping and cropping to the images, resulting in patches of size 224$\times$224. The weights of the QE are fixed, and only the MLP is adjusted. We use the AdamW optimizer with weight decay of 0.01 and batch size of 64. The learning rate starts at 0.0007 and decays using the cosine annealing strategy. Training on the LIVE, CSIQ, and CLIVE datasets lasts for 100 epochs, while for other datasets, it extends to 200 epochs.
During the testing phase, we extract five patches from each image (four corners and a center patch) and predict the quality score for each patch. These scores are then averaged to obtain the final quality score for the image. Following the scheme of~\cite{golestaneh2022no,su2020blindly}, we randomly split the dataset into 80\% training set and 20\% testing set.
For synthetically distorted datasets, we ensure that the train-test data split is based on reference images to prevent content overlap. To reduce randomness in the split, we conduct the experiments 10 times for each dataset and report the median results 
for SRCC and PLCC.

\subsection{Comparison with SOTA NR-IQA Methods}
We report the performance of SOTA NR-IQA methods in \cref{tab:perform}. 
In general, our approach has achieved SOTA performance when compared to traditional~\cite{mittal2012no,zhang2015feature,zeng2017probabilistic}, CNN-based~\cite{bosse2017deep,zhang2018blind,zhu2020metaiqa,su2020blindly,madhusudana2022image,saha2023re}, and Transformer-based methods~\cite{ke2021musiq,golestaneh2022no}, \emph{solely by linear probing}. Particularly, our method outperforms existing methods on datasets such as LIVE, CSIQ, TID2013, and KADID. Additionally, it achieves nearly SOTA performance on the CLIVE, KONIQ, and FLIVE datasets.

The DSMix-based method is easily integrated to other CNN-based IQA methods.
As shown in \cref{tab:replace}, we reproduced the results on the CSIQ, CLIVE, and KonIQ datasets based on the official open-source code of HyperIQA and TReS. 
It can be seen that DSMix has the potential to further improve the performance of these methods, showcasing strong generalization ability.

We conducted cross-dataset evaluations to demonstrate the robustness of the learned representation in our proposed method, where training and testing were conducted on different datasets. We followed the cross-database evaluation strategy of \cite{saha2023re} and the cross-database performance of all four models was evaluated using the SRCC metric.
The results presented in \cref{tab:cross} indicate that our proposed method achieves the best performance among the NR-IQA models across both synthetic and authentic distortions, demonstrating the effectiveness of our approach.

\subsection{Ablation Studies}
\noindent
\textbf{Ablation on the impact of data amount.}
To validate the effectiveness of DSMix pre-training, we employ KADIS datasets at 25\%, 50\%, and 100\% ratios. As shown in \cref{tab:amount}, the larger the quantity utilized during the pre-training process, the better the performance on the downstream IQA dataset. 
Surprisingly, our method achieves almost SOTA performance using only 50\% of the data.
Since our approach employs generated synthetic distortion data, theoretically an arbitrary amount of data can be generated for pre-training purposes.



\begin{table}[ht]
    \centering
    \begin{minipage}[t]{0.45\linewidth}
	\centering
	\caption{Performance on different amounts of synthetic data.}
	\label{tab:amount}
	\scalebox{0.85}{
		\begin{tabular}{c|cc|cc} 
			\toprule[1.3pt]
			\multirow{2}{*}{\# of KADIS} & \multicolumn{2}{c|}{CSIQ} & \multicolumn{2}{c}{CLIVE}  \\
			& SRCC & PLCC                                             & SRCC & PLCC                \\ 
			\midrule
			25\%    & 0.913 & 0.915    &0.811   &0.823                     \\
			50\%    & 0.948 & 0.951    &0.848   &0.855                    \\
			100\%   & 0.957 & 0.962    &0.873   &0.876                     \\
			\bottomrule[1.3pt]
	\end{tabular}}

    \end{minipage}
    \hfill 
    \begin{minipage}[t]{0.45\linewidth}
	\centering
	\caption{Ablation of different components.}
	\label{tab:components}
        \scalebox{0.85}{
	\begin{tabular}{c|cc|cc} 
		\toprule[1.3pt]
		\#&DSMix & KD & SRCC & PLCC  \\ 
		\midrule
		1&        &            &0.773   &0.766   \\
		2 &\checkmark 	&                &0.824   &0.837   \\
		3 &\checkmark   	& \checkmark &0.873   &0.876   \\
		\bottomrule[1.3pt]
	\end{tabular}}

    \end{minipage}
\end{table}

\begin{table}[ht]
    \centering
    \begin{minipage}[t]{0.45\linewidth}
	\centering
	\caption{Performance of different encoders on CSIQ and CLIVE datasets}
	\label{tab:encoder}
	\scalebox{0.85}{
		\begin{tabular}{c|c|cc|cc} 
			\toprule[1.3pt]
			\multirow{2}{*}{Encoder} & \multirow{2}{*}{Params} & \multicolumn{2}{c|}{CSIQ} & \multicolumn{2}{c}{CLIVE}  \\
			&                         & SRCC & PLCC                                             & SRCC & PLCC                \\ 
			\midrule
			ResNet-18                & 11.69M                  
			&0.926  &0.935  &0.831   &0.835                     \\
			ResNet-34                & 21.80M                   
			&0.944  &0.950  &0.857   &0.866                     \\
			ResNet-50                & 25.56M                   
			&0.957  &0.962  &0.873   &0.876                     \\
			\bottomrule[1.3pt]
	\end{tabular}}
    \end{minipage}
    \hfill 
    \begin{minipage}[t]{0.45\linewidth}
	\centering
	\caption{Ablation of different knowledge distillation losses.}
	\label{tab:kdloss}
	\scalebox{0.85}{
		\begin{tabular}{c|cc|cc} 
			\toprule[1.3pt]
			\#&$\mathcal{L}_{FR}$ & $\mathcal{L}_{CD}$ & SRCC & PLCC  \\ 
			\midrule
			1 &             &            &0.933   &0.942   \\
			2 &\checkmark 	&            &0.941   &0.948   \\
			3 &          	& \checkmark &0.950   &0.955   \\
			4 &\checkmark   & \checkmark &0.957   &0.962   \\
			\bottomrule[1.3pt]
	\end{tabular}}

    \end{minipage}
\end{table}



\noindent
\textbf{The impact of different components.}
Table \ref{tab:components} provides ablation experiments with respect to DSMix and knowledge distillation (KD) on CLIVE dataset. Here, \#1 represents the baseline, indicating results without using any of the components. As shown in \#2, incorporating DSMix leads to a noticeable improvement in model performance on the CLIVE dataset (+0.051 SRCC, +0.071 PLCC). Furthermore, when incorporating KD, as seen in \#3, the model achieves the best performance (+0.049 SRCC, +0.039 PLCC).

\noindent
\textbf{Ablation on the impact of encoder.}
As shown in \cref{tab:encoder}, we have selected different encoders to assess their differences in feature modeling. It can be observed that increasing the model parameters can improve performance on the CSIQ and CLIVE datasets. However, this also significantly extends the training time. Considering the trade-off, we choose to use ResNet-50. For pre-training larger models, we will explore this in future work.

\begin{figure*}
	\centering
	\begin{minipage}[b]{0.28\linewidth}
		\centering
		\subfloat[][CONTRIQUE]{\includegraphics[scale=0.32]{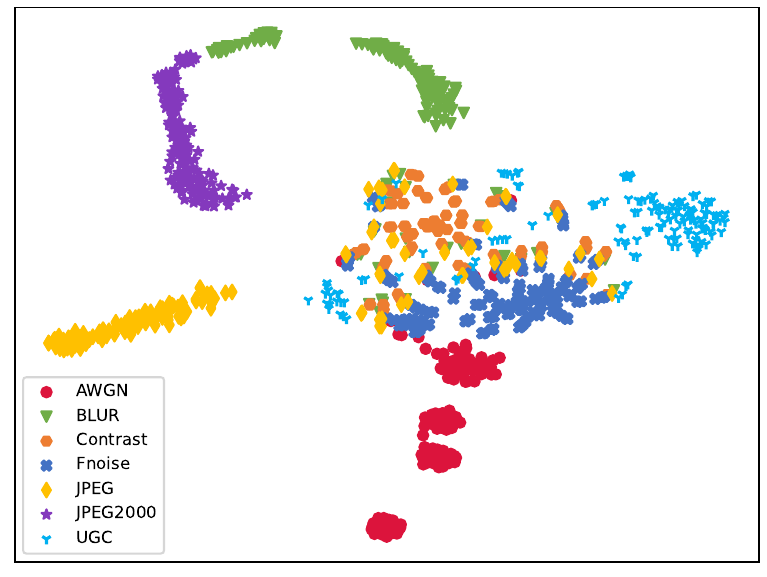}}
	\end{minipage}
	\hfill
	\begin{minipage}[b]{0.28\linewidth}
		\centering
		\subfloat[][DSMix]{\includegraphics[scale=0.32]{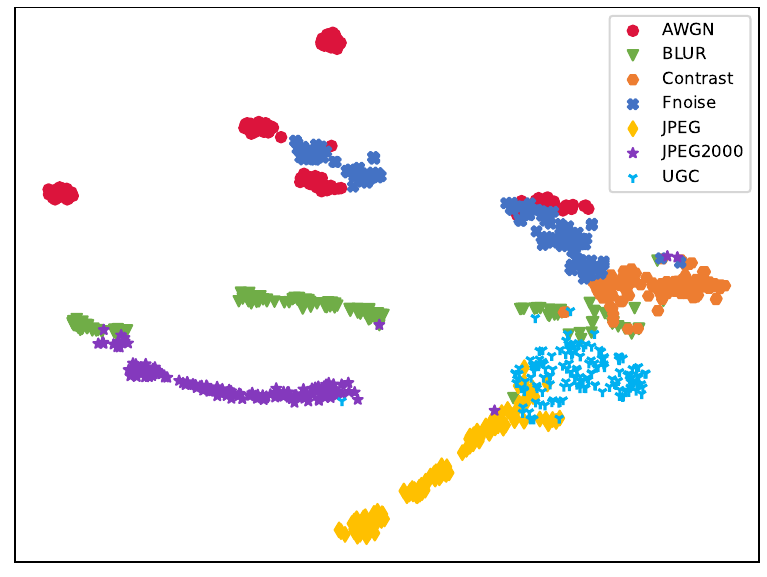}}
	\end{minipage}
	\hfill
	\begin{minipage}[b]{0.1\linewidth}
		\centering
		\subfloat[][AWGN-C]{\includegraphics[scale=0.11]{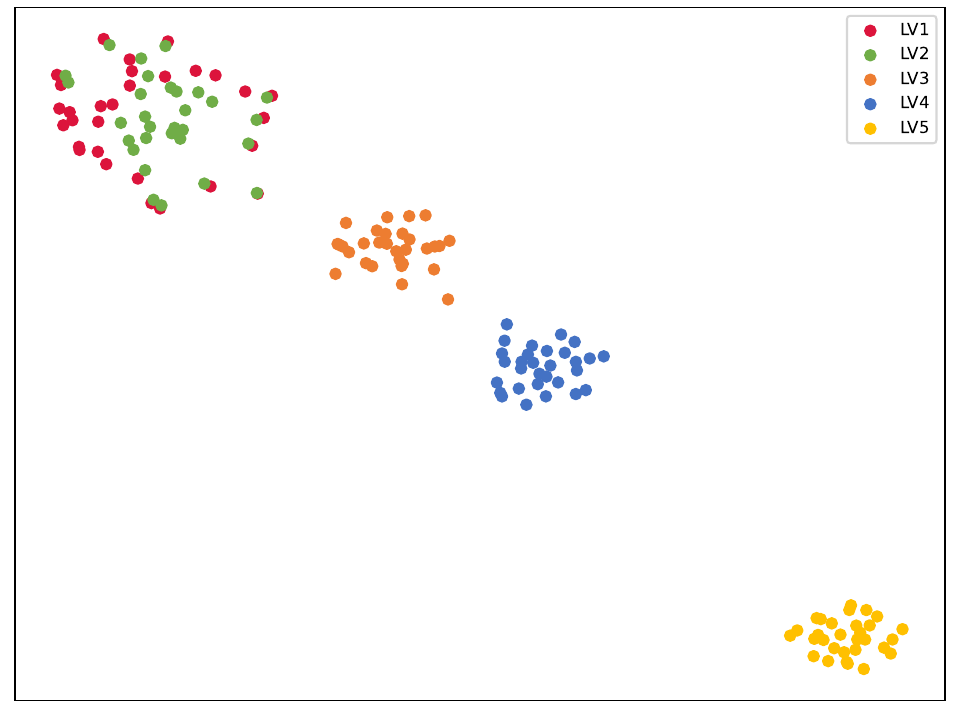}} \\
		\subfloat[][AWGN-D]{\includegraphics[scale=0.11]{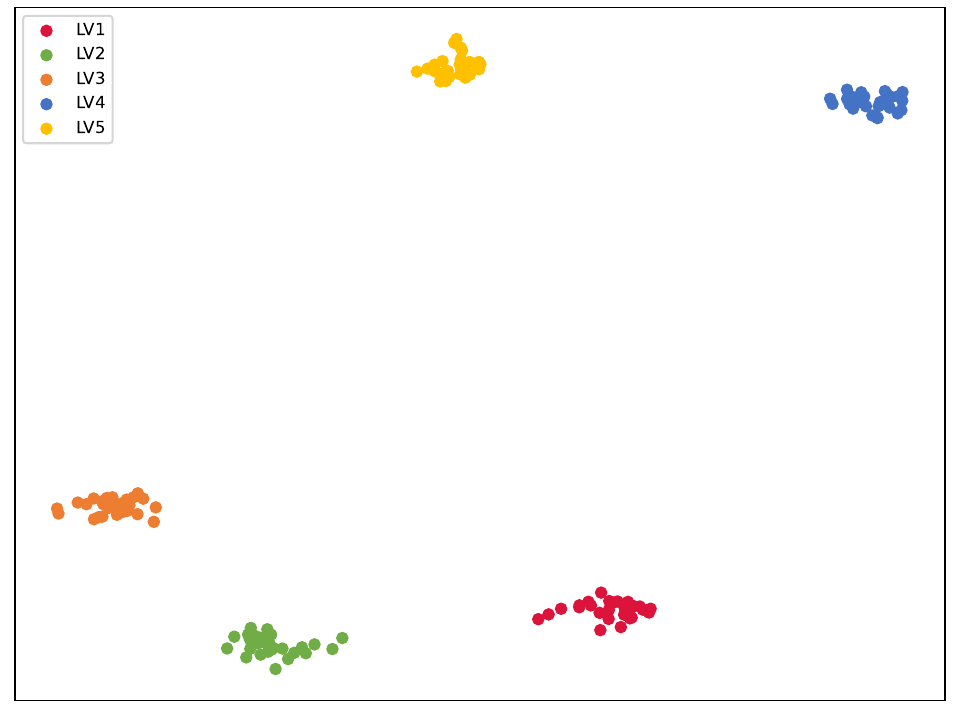}}
	\end{minipage}
	\hfill
	\begin{minipage}[b]{0.1\linewidth}
		\centering
		\subfloat[][Fnoise-C]{\includegraphics[scale=0.11]{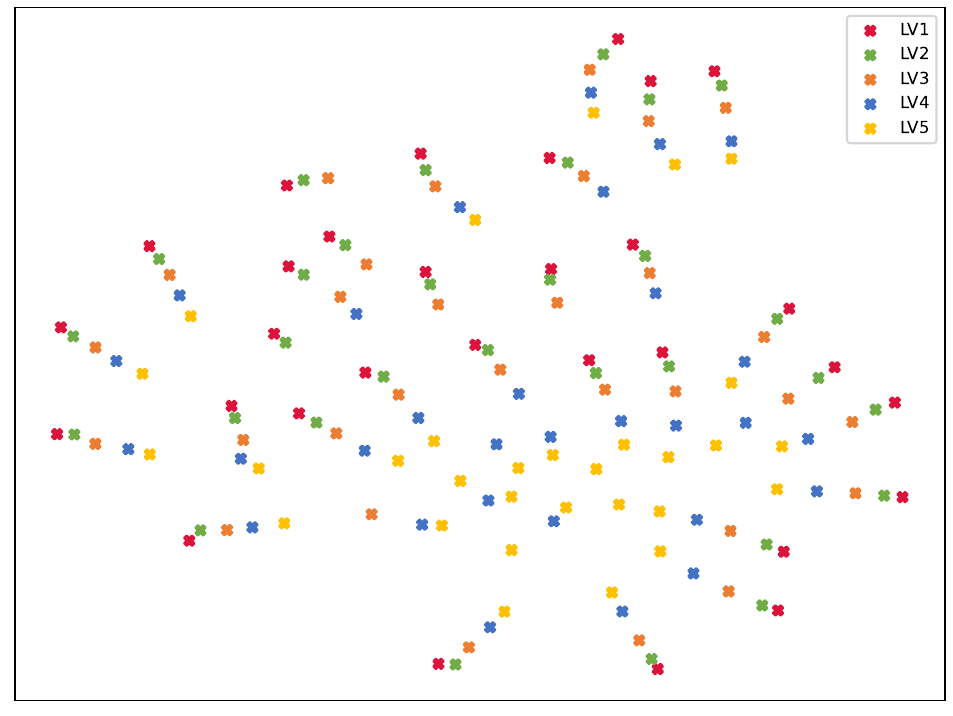}} \\
		\subfloat[][Fnoise-D]{\includegraphics[scale=0.11]{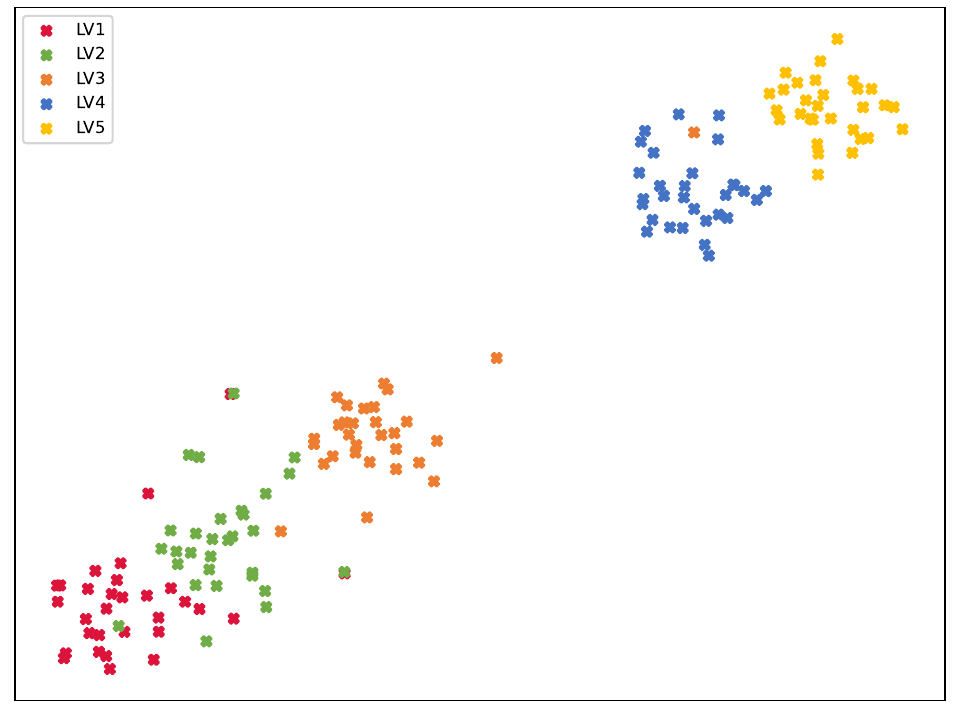}}
	\end{minipage}
	\caption{Comparison for 2D t-SNE visualization of learned representations between CONTRIQUE (``-C'') and DSMix (``-D''). Zooming in for a better view.}
	\label{fig:tsne}
\end{figure*}

\begin{figure}[h!]
	\centering
	\includegraphics[width=0.85\textwidth]{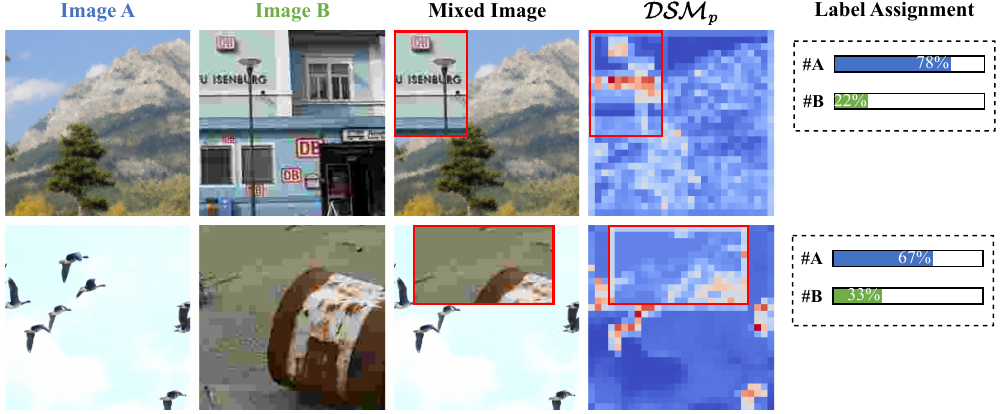}
	\caption{The visualization including image A, image B, mixed image, $\mathcal{DSM}_p$ obtained from $\mathcal{G}_\theta(\cdot )$ when inputting mixed image, and corresponding label assignments.}
	\label{fig:label}
\end{figure}

\noindent
\textbf{Ablation on the knowledge distillation loss.}
\cref{tab:kdloss} provides ablation experiments about $\mathcal{L}_{FR}$ and $\mathcal{L}_{CD}$ on CSIQ dataset. \#1 indicating results without using any knowledge distillation. 
Compared to \#1, \#2 represents the addition of feature reconstruction loss $\mathcal{L}_{FR}$, resulting in an improvement of SRCC (+0.008) and PLCC (+0.006). When cosine distance loss $\mathcal{L}_{CD}$ is added, as shown in \#3, there is a further increase in SRCC (+0.017) and PLCC (+0.013). The best performance is achieved when both losses are added.

\noindent
Additional ablations are available in the supplementary material.

\subsection{Visualizations}
\noindent
\textbf{T-SNE.}
\cref{fig:tsne} presents the 2D t-SNE~\cite{van2008visualizing} visualizations of DSMix and CONTRIQUE on the CSIQ (Synthetic-\#866) and samples from KonIQ (UGC-\#150) datasets. From (a), it is evident that CONTRIQUE lacks the ability to discern Contrast distortion, JPEG distortion, and UGC images, whereas our proposed method (b) distinctively separates them. 
The degradation level discrepancy on AWGN distortion, shown in (c) and (d), cannot be distinguished by CONTRIQUE for the first and second levels of distortion, whereas our method can. From (e) and (f), it can be observed that CONTRIQUE completely fails to differentiate Fnoise, whereas our method can still discriminate them. In conclusion, the DSMix-based pre-trained model we propose exhibits excellent feature modeling capabilities.


\noindent
\textbf{Label Assignment.}
We provide the visualization of DSMix method with Mix=2. From \cref{fig:label}, it is clear that utilizing $\mathcal{DSM}$ for label weight allocation maximizes the model's capacity to identify distinct types of distortions while further distinguishing their respective degrees.


\vspace{-3mm}
\section{Conclusion}
\vspace{-2mm}
In this paper, we propose the DSMix, a data augmentation method tailored for IQA to address the challenge of limited annotated data. By leveraging distortion-induced sensitivity maps, we quantify distortion levels, previously difficult in self-supervised learning. Additionally, we integrate semantic information from ImageNet into our model via knowledge distillation. Experimental results on multiple IQA datasets demonstrate that our method achieves SOTA performance and exhibits strong generalization ability, easily extendable to other advanced IQA models. 

\vspace{-3mm}
\section*{Acknowledgements}
\vspace{-2mm}
This work was partially supported by the National Natural Science Foundation of China (No. 62272227 \& No. 62276129), and the Natural Science Foundation of Jiangsu Province (No. BK20220890).

\bibliographystyle{splncs04}
\bibliography{main}

\end{document}



\section*{Appendix}

\begin{figure}[hb]
	\centering
	\includegraphics[width=\textwidth]{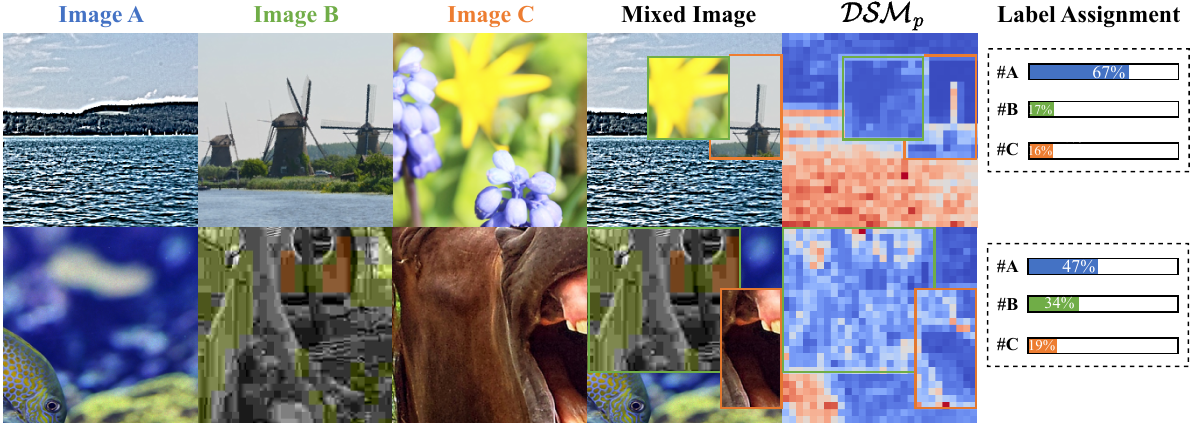}
	\captionof{figure}{The visualization including image A, image B, image C and mixed image, $\mathcal{DSM}_p$ obtained from $\mathcal{G}_\theta(\cdot )$ when inputting mixed image, and corresponding label assignments.}
	\label{fig:sup-label}
\end{figure}

\section{Comparison with popular data augmentation methods}
\begin{figure}[h]
	\centering
	\includegraphics[width=0.9\textwidth]{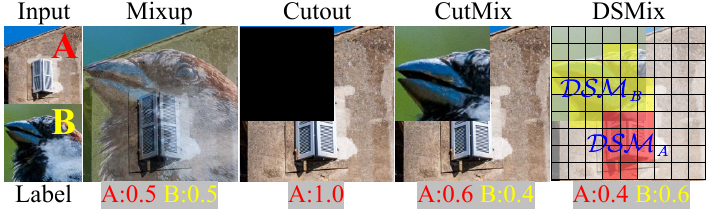}
	\caption{Existing mixup-based data augmentation methods. A and B represent different types of distortion.}
	\label{fig:cmp}\
\end{figure}

Previous mixup-based data augmentation methods towards classification have a common underlying assumption that the linear interpolation ratio of the mixed label should remain consistent with the ratio proposed in the input interpolation. This is exemplified by Mixup~\cite{zhang2017mixup}, Cutout~\cite{devries2017improved}, and CutMix~\cite{yun2019cutmix}, as shown in Fig.~\ref{fig:cmp}. However, in the context of IQA, this can lead to a phenomenon where there is no effective distortion information in the mixed image during the augmentation process, yet there still exists a response in the label space. Additionally, as depicted in Fig.~\ref{fig:cmp}, pixels in the background contribute less to the label space compared to pixels in those salient regions. However, traditional mixup-based methods fail to effectively differentiate such pixels.

To address \emph{the gap between input and mixed labels}, we propose DSMix, which is based on the Distortion-Induced Sensitivity Map for generating mixed labels. Here, the confidence of the label increases with the distortion degree in the input image. To the best of our knowledge, this is the \emph{first work} specifically tailored for data augmentation in the IQA domain. Through pre-training using DSMix, our model achieves state-of-the-art performance without the need for any fine-tuning, solely relying on linear probing.

\section{Computational Performance Comparison}
\begin{figure}[t]
	\centering
	\includegraphics[width=0.8\textwidth]{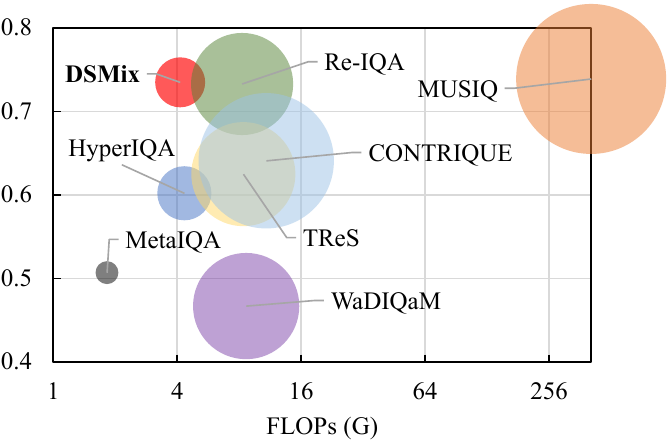}
	\caption{Comparison of PLCC performance and FLOPs for different methods on the FLIVE dataset. The radius of the circles represents the number of model parameters.}
	\label{fig:scatter}
\end{figure}

Fig.~\ref{fig:scatter} demonstrates the performance w.r.t. the required computational resources. FLOPs are used to quantify the computational cost, and the number of parameters is represented by the size of the circles. It is evident that the proposed method achieves higher performance with fewer resources (such as FLOPs and parameters) compared to other methods. Note that DSMix exhibits superior performance compared to state-of-the-art methods with extremely low computational resources, demonstrating its high practicality for real-world applications.

\section{Evaluation Criteria}
SRCC is defined as follows:
\begin{equation}
	\mathrm{SRCC}=1-\frac{6 \sum_{i=1}^{n} d_{i}^{2}}{n\left(n^{2}-1\right)}
\end{equation}
where $n$ is the number of test images and $d_{i}$ denotes the difference between the ranks of $i$-th test image in ground-truth and the predicted quality scores. PLCC is defined as:
\begin{equation}
	\mathrm{PLCC}=\frac{\sum_{i=1}^{n}\left(u_{i}-\bar{u}\right)\left(v_{i}-\bar{v}\right)}{\sqrt{\sum_{i=1}^{n}\left(u_{i}-\bar{u}\right)^{2}} \sqrt{\sum_{i=1}^{n}\left(v_{i}-\bar{v}\right)^{2}}}
\end{equation}
where $u_{i}$ and $v_{i}$ denote the ground-truth and predicted quality scores of the $i$-th image, and $\bar{u}$ and $\bar{v}$ are their mean values, respectively.

\section{More ablation study results}
\subsection{Ablation study on DSM pre-training}
\cref{tab-dsm} presents the ablation study results for the first stage, comparing the use of ground truth error maps and gradient maps. It can be seen that the model yields very similar results when using GT DSM and $\mathcal{DSM}_p$. 
Using $\mathcal{DSM}_p$ can make our proposed pre-training method a true \emph{blind} quality model. 
\begin{table}[h]
	\small
	\centering
	\caption{Ablation of DSM ($\mathcal{DSM}_p$: predicted DSM)}
	\label{tab-dsm}
	\scalebox{1.0}{
		\begin{tabular}{@{}c|cc|cc|cc@{}}
			\toprule[1.3pt]
			& \multicolumn{2}{c|}{CSIQ} & \multicolumn{2}{c|}{CLIVE} &\multicolumn{2}{c}{LIVE} \\
			& SRCC        & PLCC        & SRCC        & PLCC     & SRCC   & PLCC   \\ \midrule
			Gradient map & 0.934       & 0.941       & 0.844       & 0.848    & 0.956  & 0.962    \\
			$\mathcal{DSM}_p$          & 0.957       & 0.962       & 0.873    & 0.883  & 0.974  & 0.974    \\
			GT-DSM       & \textbf{0.961}       & \textbf{0.964}       & -       & -   & 0.976  & 0.977 \\ \bottomrule[1.3pt]
	\end{tabular}}
\end{table}

\subsection{Ablation study on knowledge distillation}
Apart from the CLIVE results mentioned in paper, we also present ablation results on the LIVE, CSIQ, and KonIQ datasets without using KD in \cref{tab-kd} below. 
It is worth noting that even without employing KD, our proposed method achieves performance close to the SOTA performance through linear probing. 
\begin{table}[h]
	\small
	\centering
	\caption{Ablation of KD}
	\label{tab-kd}
	\scalebox{1.0}{
		\begin{tabular}{@{}c|cccccc@{}}
			\toprule[1.3pt]
			& \multicolumn{2}{c}{LIVE} & \multicolumn{2}{c}{CSIQ} & \multicolumn{2}{c}{KonIQ} \\
			& SRCC        & PLCC       & SRCC        & PLCC       & SRCC        & PLCC        \\ \midrule
			DSMix w/o KD & 0.970        & 0.972      & 0.952       & 0.960       & 0.903       & 0.914       \\
			DSMix w/ KD & 0.974        & 0.974      & 0.957       & 0.962       & 0.915       & 0.925       \\
			\bottomrule[1.3pt]
	\end{tabular}}
\end{table}

\subsection{Ablation study on linear probing}
\cref{tab-lp} presents the ablation study results for the third stage, comparing the use of ridge regression (RR), full fine-tuning (FF), linear probing (LP). It can be observed that the FF manner can further improve the model's performance.
\begin{table}[h]
	\small
	\centering
	\caption{Ablation of LP}
	\label{tab-lp}
	\scalebox{1.0}{
		\begin{tabular}{@{}c|cc|cc|cc@{}}
			\toprule[1.3pt]
			& \multicolumn{2}{c|}{CSIQ} & \multicolumn{2}{c|}{CLIVE} & \multicolumn{2}{c}{LIVE} \\
			& SRCC        & PLCC        & SRCC        & PLCC    & SRCC   & PLCC    \\ \midrule
			RR           & 0.953       & 0.956       & 0.864       & 0.870   & 0.972  & 0.973   \\
			LP           & 0.957       & 0.962       & 0.873       & 0.883   & 0.974  & 0.974   \\
			FF           & \textbf{0.963}       & \textbf{0.970}       & \textbf{0.891}       & \textbf{0.907}    & \textbf{0.980}       & \textbf{0.981}     \\ \bottomrule[1.3pt]
	\end{tabular}}
\end{table}

\subsection{Ablation study on patch size}
Tab.~\ref{tab:psize} presents the impact of patch size on the final quality assessment during different DSMix data augmentation processes. It is evident that as the patch size decreases, the model's evaluation performance gradually improves. Therefore, we have opted for a patch size of 8; going any further lower would result in excessively high computational complexity for the model.

\begin{table}[h]
\centering
\caption{Ablation study on patch size.}
\label{tab:psize}
\scalebox{1.0}{
\begin{tabular}{c|cc|cc|cc}
\toprule[1.2pt]
\multirow{2}{*}{\begin{tabular}[c]{@{}c@{}}Patch\\ size\end{tabular}} & \multicolumn{2}{c|}{LIVE} & \multicolumn{2}{c|}{CSIQ} & \multicolumn{2}{c}{LIVEC} \\
     & SRCC    & PLCC     & SRCC      & PLCC      & SRCC      & PLCC  \\ \midrule
32   & 0.950   & 0.952    & 0.934     & 0.940     & 0.849     & 0.855   \\
16   & 0.963   & 0.965    & 0.943     & 0.951     & 0.860     & 0.864   \\
8    & 0.974   & 0.974    & 0.957     & 0.962     & 0.873     & 0.883   \\   \bottomrule[1.2pt]
\end{tabular}}
\end{table}

\section{Comparison with other SOTA methods}

\cref{tab-comp} presents a comparison with other SOTA methods. CLIP-IQA+ \cite{wang2023exploring} is pretrained on the KonIQ and tested on other datasets, LIQE \cite{zhang2023blind} involves end-to-end training by combining  the CLIP model for multitask learning, and MANIQA \cite{yang2022maniqa} is trained from scratch. It is evident that our proposed method achieves close to SOTA performance \textit{solely by linear probing} without any assistance, \eg, language.
\begin{table}[h]
\small
\centering
\caption{Comparison with SOTA methods.}
\label{tab-comp}
\scalebox{1.0}{
	\begin{tabular}{@{}c|cc|cc|cc@{}}
		\toprule[1.3pt]
		& \multicolumn{2}{c|}{KADID}      & \multicolumn{2}{c|}{CLIVE}     & \multicolumn{2}{c}{KonIQ}       \\
		& SRCC           & PLCC           & SRCC           & PLCC          & SRCC           & PLCC           \\ \midrule
		CLIP-IQA+ & -              & -              & 0.805          & 0.832         & 0.895          & 0.909          \\
		LIQE      & 0.930          & 0.931          & \textbf{0.904} & \textbf{0.910} & \textbf{0.919} & 0.908          \\
		MANIQA    & \textbf{0.944} & \textbf{0.946} & 0.871          & \underline{0.887}    & 0.880          & \underline{0.915}          \\ \midrule
		Ours      & \underline{0.943}     & \underline{0.945}     & \underline{0.873}     & 0.883         & \underline{0.915}    & \textbf{0.925}  \\ \bottomrule[1.3pt]
\end{tabular}}
\end{table}

\section{Extended Visualization Results}
\subsection{Label Assignment}
\cref{fig:sup-label} provides the visualization of label assignment when Mix=3. Despite the complexity and diversity of the inputs, $\mathcal{DSM}_p$ is still capable of effectively capturing various types of distortions and degradation levels. This ensures the accuracy of label weight allocation.

\begin{figure}[h]
	\centering
	\includegraphics[width=0.85\textwidth]{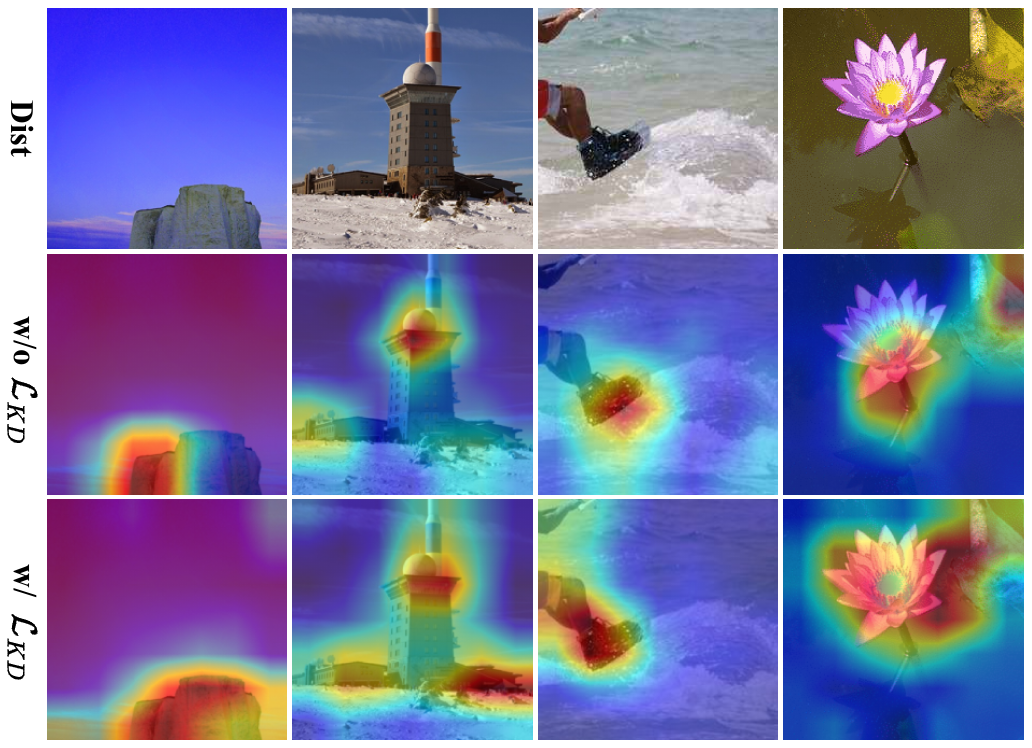}
	\caption{The visualization of Grad-CAM for knowledge distillation loss $\mathcal{L}_{KD}$. Where Dist denotes distorted images.}
	\label{fig:sup-kd}
\end{figure}

\subsection{Grad-CAM}
\cref{fig:sup-kd} provides visualizations of the ablation study on $\mathcal{L}_{KD}$, and it is evident that with $\mathcal{L}_{KD}$, the quality encoder is capable of simultaneously attending to distortion information and incorporating the semantic content of the image. This results in a better alignment with human perceptual evaluation.

\begin{figure*}
	\centering
	\includegraphics[width=0.995\textwidth]{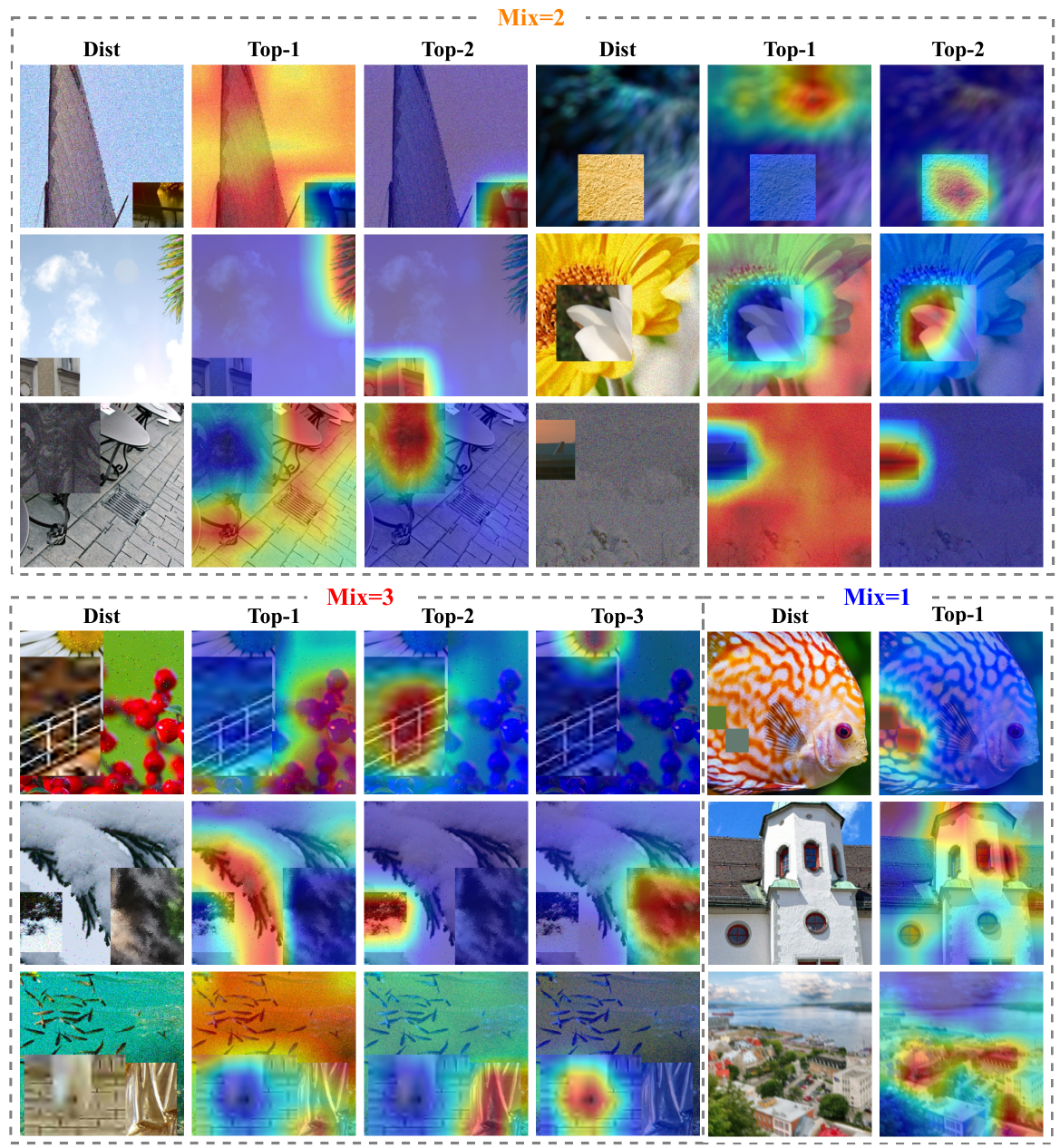}
	\caption{Grad-CAM visualizations on augmented images.}
	\label{fig:sup-cam}
\end{figure*}

\begin{figure*}
	\centering
	\includegraphics[width=0.995\textwidth]{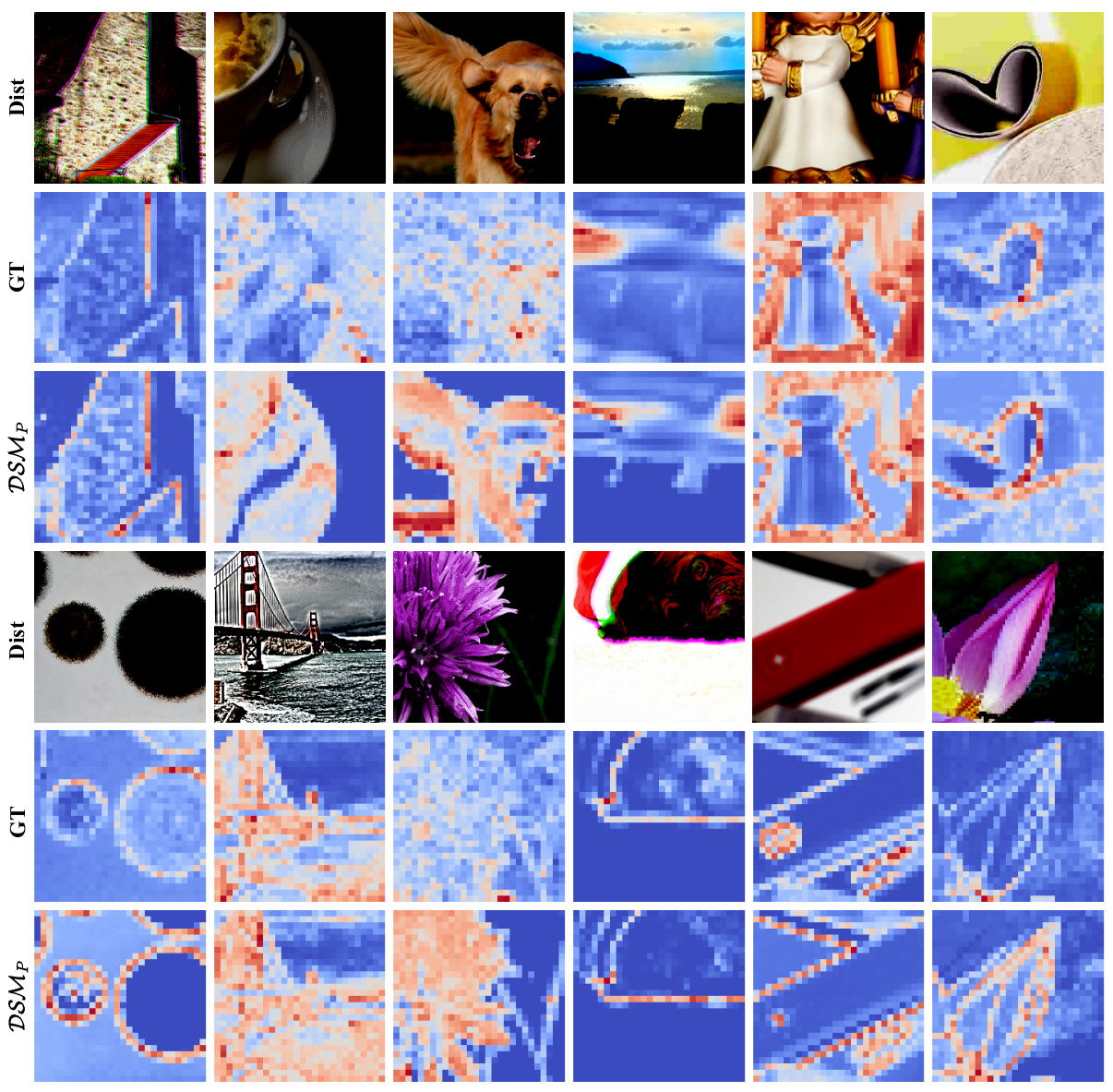}
	\caption{Visualizations of predicted DSM. Dist denotes distorted images randomly sampled from KADIS.}
	\label{fig:sup-dsm}
\end{figure*}

\cref{fig:sup-cam} provides visualizations of Grad-CAM results with Mix=1, 2, and 3. It can be obviously seen that our proposed models are capable of effectively capturing diverse types of distortions.

\subsection{Predicted DSM}
\cref{fig:sup-dsm} shows the visualization of the output $\mathcal{DSM}_p$ from model $\mathcal{G}_\theta(\cdot )$ and the ground-truth (GT). It can be observed that $\mathcal{DSM}_p$ accurately extracts the salient distorted regions in the image, even outperforming the given GT in some cases (\eg, the dog and flower in the third column). This demonstrates the effectiveness of $\mathcal{DSM}$.

%
%
\bibliographystyle{splncs04}
\bibliography{main}